\documentclass{article} 

\usepackage[table]{xcolor}
\usepackage{graphicx} 
\usepackage{float}
\usepackage{wrapfig}

\usepackage{algorithm}
\usepackage{algorithmic}
\usepackage{amsmath}
\def\bx{{\bf x}}
\def\by{{\bf y}}
\def\bby{{\bf \bar{y}}}

\def\bw{{\bf w}}

\def\argmax{\text{argmax}}

\def \nn {\nonumber} 
\def \nnnl {\nonumber \\}

\newcommand{\beqn}{\begin{equation}}
\newcommand{\eeqn}{\end{equation}}
\newcommand{\bitm}{\begin{itemize}}
\newcommand{\eitm}{\end{itemize}}
\newcommand{\bfig}{\begin{figure}}
\newcommand{\efig}{\end{figure}}
\newcommand{\baln}{\begin{align}}
\newcommand{\ealn}{\end{align}}
\newcommand{\btblr}{\begin{tabular}}
\newcommand{\etblr}{\end{tabular}}
\newcommand{\btbl}{\begin{table}}
\newcommand{\etbl}{\end{table}}
\newcommand{\bsec}{\begin{section}}
\newcommand{\esec}{\end{section}}
\newcommand{\bchr}{\begin{chapter}}
\newcommand{\echr}{\end{chapter}}
\newcommand{\bseqn}{\begin{subequations}}
\newcommand{\eesqn}{\end{subequations}}

\newcommand{\inner}[2]{ #1^\top #2}

\newcommand{\RR}{\mathbf{R}}
\newcommand{\BlackBox}{\rule{1.5ex}{1.5ex}}  
\newenvironment{proof}{\par\noindent{\bf Proof\ }}{\hfill\BlackBox\\[2mm]}

\newtheorem{theorem}{Theorem}

\usepackage{nips11submit_e,times}

\nipsfinalcopy

\title{Online Learning with Preference Feedback}

\author{
Pannagadatta K. Shivaswamy  \\
Department of Computer Science \\
 Cornell University,  Ithaca NY \\
\texttt{pannaga@cs.cornell.edu} \\
\And
Thorsten Joachims \\
Department of Computer Science \\
 Cornell University,  Ithaca NY \\
\texttt{tj@cs.cornell.edu} 
}

\begin{document} 

\maketitle
\begin{abstract}
We propose a new online learning model for learning with preference feedback. The model is especially suited for applications like web search and recommender systems, where preference data is readily available from implicit user feedback (e.g. clicks). In particular, at each time step a potentially structured object (e.g. a ranking) is presented to the user in response to a context (e.g. query), providing him or her with some unobserved amount of utility. As feedback the algorithm receives  an improved object that would have provided higher utility. We propose a learning algorithm with provable regret bounds for this online learning setting and demonstrate its effectiveness on a web-search application. The new learning model also applies to many other interactive learning problems and admits several interesting extensions.
\end{abstract}

\section{Introduction}

Our new learning model is motivated by how users interact with a web-search engine or a recommender system. At each time step, the user issues a query and the system responds by supplying a list of results. The user views some of the results and selects those that he or she prefers. Here are two such examples:
\vspace{-0.5\baselineskip}
\begin{description}\addtolength{\itemsep}{-0.45\baselineskip}
\item[Web Search:] In response to a query, the search engine presents the ranking $[A,B,C,D,E,...]$ and observes that the user clicks on documents $C$ and $D$.
\item[Movie Recommendation:] An online service recommends movie A to a user. However, the user ignores the recommendation and instead rents another movie B after some browsing.
\end{description}
In both cases the user feedback comes in the form of a preference. In the web search example, we can infer that the user would have preferred the ranking $[C,D,A,B,E,...]$ over the one we presented \cite{Joachims/etal/07a}. In the recommendation example, movie $B$ was preferred over movie $A$. The cardinal utilities of the predictions, however, are never observed, and the algorithm typically does not get the optimal ranking/movie as feedback. 

This preference feedback is different from conventional online learning models. In the simplest form of the multi-armed bandit problem \cite{ACSF02,AuerCF02,olbook}, an algorithm chooses an action (out of $K$ possible actions) and observes  reward only for that action. Conversely, rewards of all possible actions are revealed in the case of learning with expert advice \cite{olbook}. Our model, where the ordering of two arms is revealed (the one we presented and the one we receive as feedback), sits between the expert and the bandit setting. A similar relationship holds for online convex optimization  \cite{Zink03} and online convex optimization in the bandit setting \cite{FKM05}, which can be viewed as continuous extensions of the expert and the bandit problems respectively, since they rely on observing either a full convex function or the value of a convex functions after each iteration. Most closely related to our work is the dueling bandits setting \cite{Yue/etal/09a,Yue/Joachims/09a}, but existing algorithms are known to converge rather slowly.

In the following, we formally define the online preference learning model and a notion of regret, propose a simple algorithm for which we prove a regret bound, and empirically evaluate the algorithm on a web-search problem.

 
\section{Online Preference Learning Model}
The online preference learning model is defined as follows. At each round $t$, the learning algorithm receives a context  $\bx_t \in {\cal X}$ and presents a (possibly structured) object $\by_t \in {\cal Y}$. In response, the user returns an object $\bby_t \in {\cal Y}$ which the algorithm receives as feedback. For example, in web-search, a user issues a query and is presented with a ranked list of URL's ($\by_t$). The user interacts with the ranking that was provided to her by clicking on results that are relevant to her. This user interaction allows us to infer a better ranking $\bby_t$ to this user. 

We assume that the user evaluates rankings according to a utility function $U(\bx,\by)$ that is unknown to the learning algorithm. A natural way to define regret in this model is based on the difference in utility $U(\bx_t,\by^*_t)-U(\bx_t,\by_t)$ between the object $\by_t$ we present and the best possible objects $\by^*_t=\argmax_\by U(\bx_t,\by)$ that could have been presented. The goal of an algorithm is to minimize
\begin{align}
\label{eq:linregret}
&\text{REGRET}_T := \frac{1}{T} \sum_{t=1}^T \left( U(\bx_t,\by^*_t) - U(\bx_t,\by_t)  \right). 
\end{align}
To prove bounds on the regret, we specify the properties of the user's preference feedback more precisely. We say that user feedback is {\em $\alpha$-informative}, if for some $\alpha \in (0,1]$ and $\xi_t \ge 0$
\begin{align}
\label{eq:inf-feedback-relax}
& \left( U(\bx_t, \bby_t) - U(\bx_t,\by_t) \right) = \alpha \left( U(\bx_t,\by_t^*) - U(\bx_t,\by_t ) \right) - \xi_t.
\end{align}
Intuitively, the above definition describes the quality of feedback by how much the utility of the user feedback $\bby_t$ is higher than that of the algorithm's prediction $\by_t$ in terms of an (unknown) fraction $\alpha$ of the maximum possible utility range. Note that $\xi_t \ge 0$ is a slack variable that captures noise in the feedback. 

In the following, we use a linear model for the utility function
\begin{align}
\label{eq:linutil}
U(\bx,\by) = \bw^{*\top} \phi(\bx,\by),
\end{align}
where $\bw^* \in \mathbf{R}^N$ is an unknown parameter vector and $\phi: {\cal X} \times {\cal Y} \rightarrow \RR^N$ is a joint feature map such that $\|\phi(\bx,\by)\| \le R$ for any $\bx \in {\cal X}$ and $\by \in {\cal Y}$.

\floatstyle{boxed}
\restylefloat{figure}
\begin{wrapfigure}{r}{0.5\textwidth}
\begin{algorithmic}
\STATE Initialize $\bw_1 \leftarrow {\mathbf 0}$
\FOR{$t=1$ {\bf to} $T$}
\STATE Observe $\bx_t$
\STATE Present $\by_t \leftarrow \argmax_{\by \in {\cal Y}} \bw_t^\top \phi(\bx_t,\by)$
\STATE Obtain feedback $\bby_t$
\STATE Update: $\bw_{t+1} \leftarrow \bw_{t} + \phi(\bx_t,\bby_t) - \phi(\bx_t,\by_t)$
\ENDFOR 
\end{algorithmic}
\caption{\label{perceptron} Preference Perceptron.} 
\end{wrapfigure}

\floatstyle{plain}
\restylefloat{figure}

\section{Algorithm}

We propose the algorithm in Figure~\ref{perceptron} for the online preference learning problem.
It maintains a vector $\bw_t$ and predicts the object with the highest utility according to $\bw_t$ in each iteration $t$. It then receives feedback $\bby_t$ and updates $\bw_t$ in the direction $\phi(\bx_t,\bby_t) - \phi(\bx_t,\by_t)$.

\begin{theorem}
\label{thm:bound}
Under $\alpha$-informative feedback the algorithm in Figure \ref{perceptron} has regret 
\begin{align}
\label{eq:percept-reg}
\text{REGRET}_T \le \frac{1}{\alpha T} \! \sum_{t=1}^T \xi_t + \frac{2R \|\bw^* \|}{ \alpha \sqrt{T}}.
\end{align}
\end{theorem}

Proof of the above theorem is provided in the Appendix \ref{app:proof}. When the user feedback is noise free, the first term on the right hand side of the above bound vanishes. The average regret in this case approaches zero at the rate $1/\sqrt{T}$. In addition to this result, we have the following extensions which we cannot provide here due to space limitations:

\vspace{-0.5\baselineskip}
\begin{itemize}\addtolength{\itemsep}{-0.45\baselineskip}
\item{It is possible to further weaken the requirement on the feedback. Instead of requiring $\alpha$-informative feedback, the user is required to give $\alpha$-informative feedback in expectation. We can show a result similar to that in Theorem \ref{thm:bound} in this case.}
\item{It is also possible to show that an algorithm different from Algorithm~\ref{perceptron} can minimize any convex loss (under mild assumptions) defined on the utility difference $\bw^{*\top} \left(\phi(\bx_t,\by_t) - \phi(\bx_t,\by_t^*) \right) $.}
\end{itemize}

\section{Experiments}
\begin{figure*}[tbh!!!]
\begin{center}
\vskip -0.1in
\includegraphics[width=2.7in]{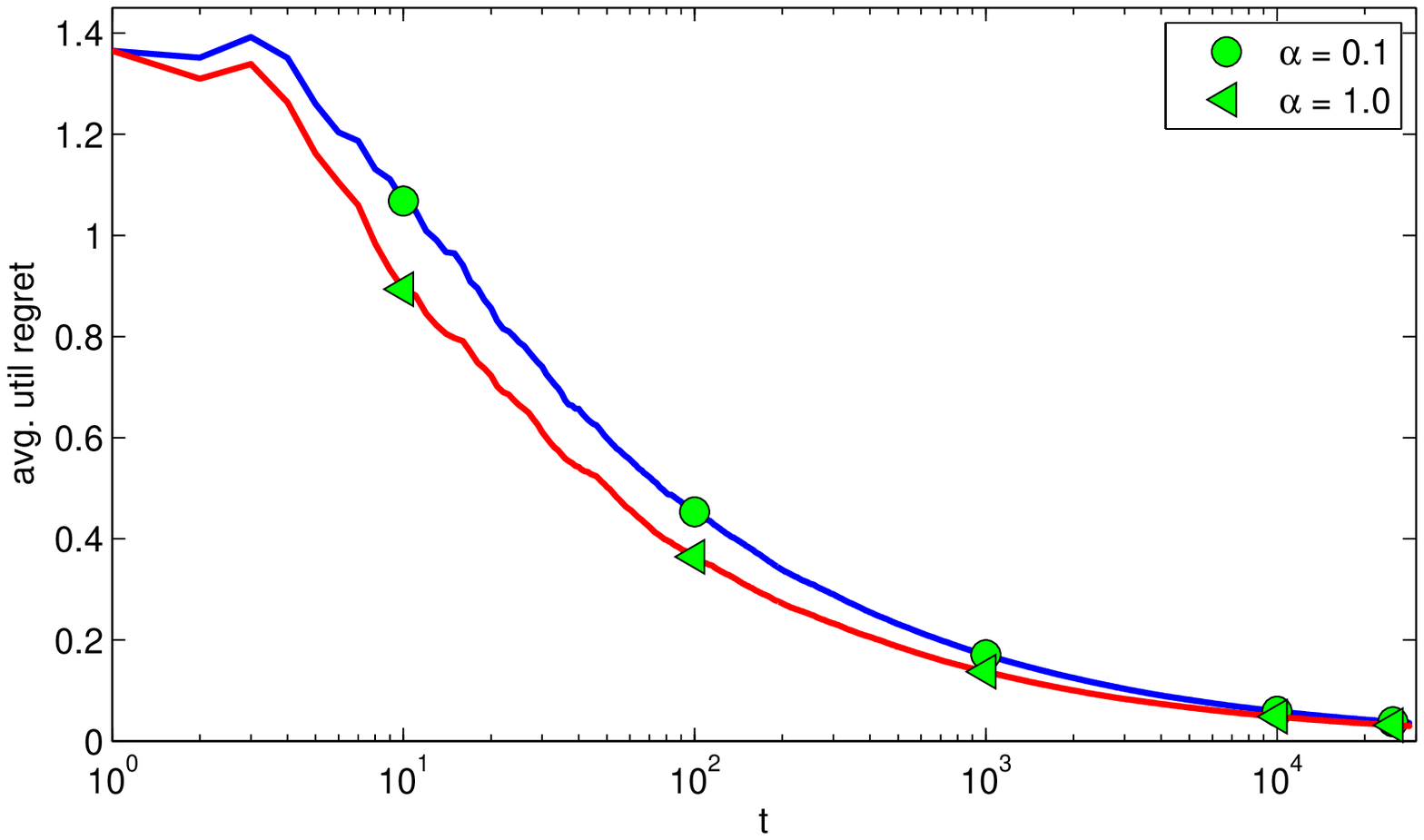}
\includegraphics[width=2.7in]{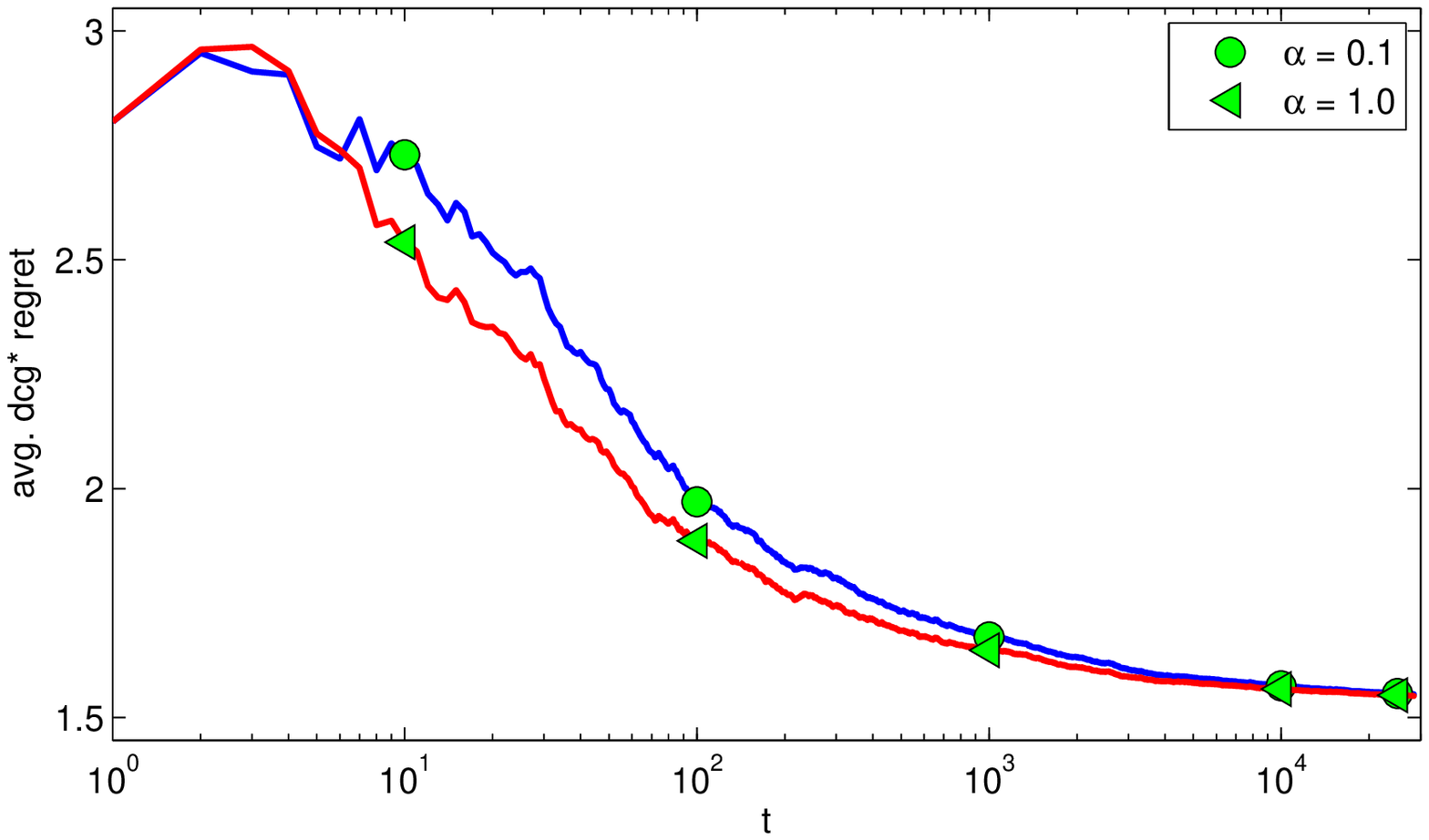}
\vskip -0.1in
\caption{\label{fig:clean-feedback} Average regret versus time based on noise free $\alpha$-informative feedback.}
\vskip -0.1in
\end{center}
\end{figure*}

We applied our Preference Perceptron algorithm to the Yahoo! learning to rank dataset \cite{ylr11}. This dataset consists of query-url features (denoted as $\bx^q_i$ for query $q$ and URL $i$ for that particular query) with a relevance rating $r^q_i$ which ranges from zero (irrelevant) to four (perfectly relevant).  We first computed the best least squares fit to the relevance labels from the features using the entire dataset and all the utilities in our experiment are reported with respect to this $\bw^*$. 

To pose ranking as a structured prediction problem, we defined our joint feature map as follows:
\begin{align}
\label{eq:ranking-utility}
\bw^\top\phi(q,\by) =  \sum_{i=1}^5 \frac{\bw^\top \bx^q_{\by_i}}{\log(i+1)}
\end{align}
In the above equation, $\by$ denotes a ranking. In particular, $\by_i$ is the index of the URL which is placed at position $i$ in the ranking. Thus, the above measure
considers the top five URLs for a query $q$ and computes a score based on a graded relevance.  The above feature-map and utility are motivated from the definition:
$\text{DCG@5}(q,\by) = \sum_{i=1}^5 \frac{r^q_{\by_i}}{\log(i+1)}.$ Effectively, our utility score \eqref{eq:ranking-utility} mimics DCG@5 by replacing the relevance label with a linear prediction based on the features. 

For query $q_t$ at time step $t$, the Preference Perceptron algorithm present the ranking $\by^q_t$ that maximizes $\bw_t^\top \phi(q,\by)$. Note that this merely amounts to sorting documents by the scores $\bw_t^\top \bx^{q_t}_i$, which can be done very efficiently. Once a ranking ($\by^{q_t}$) was presented to a user, the user returns a ranking $\bby^{q_t}$. The exact nature of user feedback differed in the two experiments; the details of feedback can be found below.  Query ordering was randomly permuted twenty times and all the results reported  are an average over the runs.

The utility regret in Eqn. \eqref{eq:linregret}, based on the definition of utility in \eqref{eq:ranking-utility}, is given by $\frac{1}{T} \sum_{t=1}^T(\bw^{*\top} \phi(q_t,\by^{q_t*}) - \phi(q_t,\by^{q_t})) $. Here $\by^{q_t*}$ denotes the optimal ranking with respect to $\bw^*$. We also present our results on another quantity which we refer to as the {\bf DCG* regret}. Since for every query-URL pair there is a manual relevance judgment in the dataset, optimal DCG can be computed by sorting the relevance score. In DCG* regret, we measure the difference between the DCG of the optimal ranking and that of the rankings we present in each step.

\paragraph{$\alpha$-informative feedback}
The goal of the first experiment was to see how the regret of the algorithm changes with $\alpha$, assuming $\alpha$-informative feedback without noise. Once a ranking was presented, the feedback was obtained as follows:  given a ranked list, the simulated user would go down the list and would stop when she found  five URL's such that,  when they are placed at the top of the list (in the order of their utilities),  gave noise free $\alpha$-informative feedback (i.e. $\xi_t=0$) based on $\bw^*$.   Figure \ref{fig:clean-feedback} shows the results for this experiment for two different $\alpha$ values. As expected, the regret with $\alpha=1.0$ is lower compared to the regret with respect $\alpha=0.1$. Note, however, that the difference between the two curves is much smaller than a factor of ten. This is because, strictly $\alpha$-informative feedback is also strictly $\beta$-informative feedback for any $\beta \le \alpha$. So, there could be several instances where user feedback was much stronger than what was required. Since the slack variables are zero, the average utility regret approaches zero as expected. 



\begin{figure*}[tbh!!]
\begin{center}
\vskip -0.1in
\includegraphics[width=2.7in]{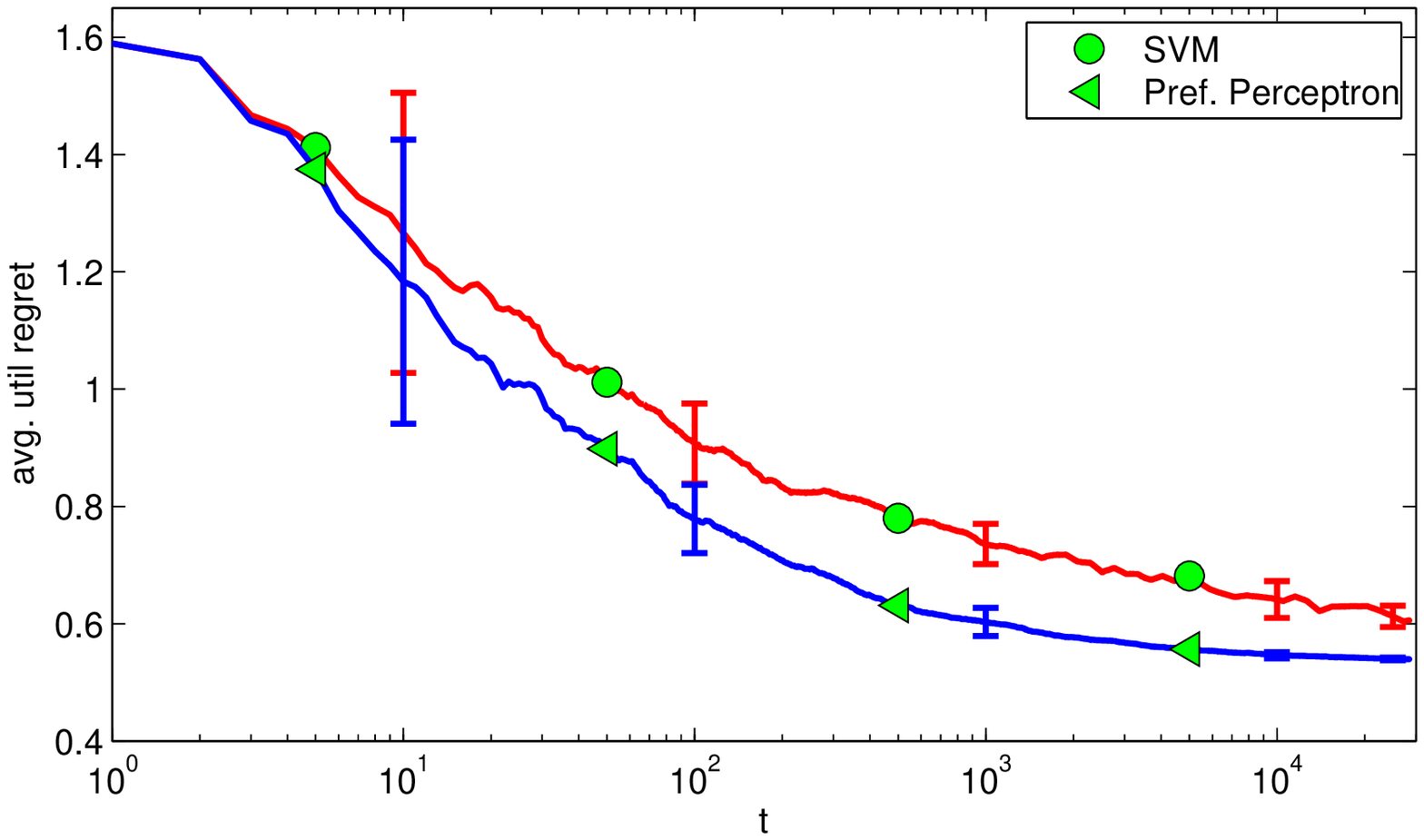}
\includegraphics[width=2.7in]{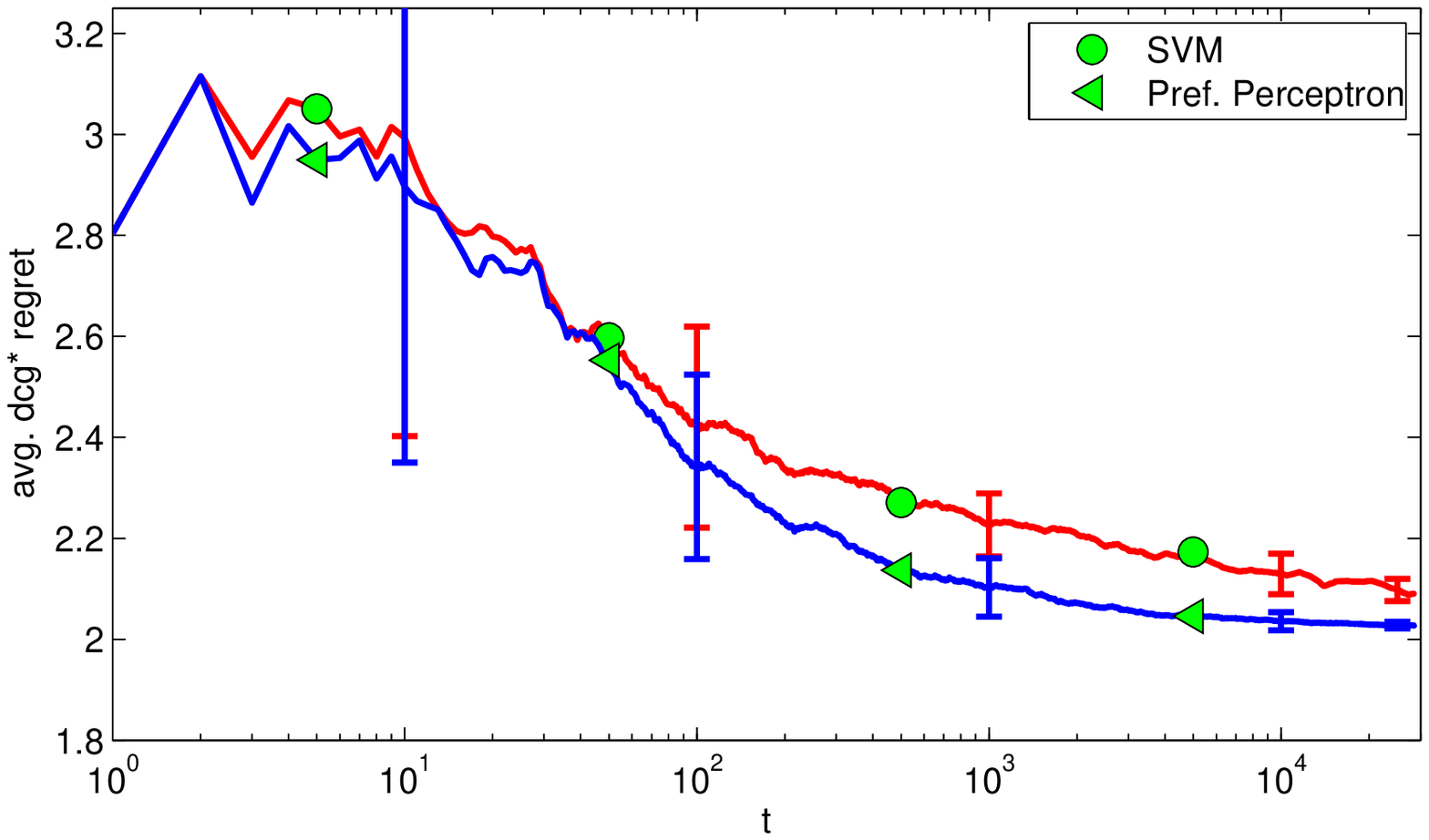}
\vskip -0.1in
\caption{\label{fig:rel-feedback} Regret versus time based on actual relevance labels.}
\vskip -0.1in
\end{center}
\end{figure*}

\paragraph{Relevance label feedback}
In this experiment, feedback was based on the actual relevance labels in the dataset as follows: given a ranking for a query, the user would go down the list inspecting the top 25 (or all the URLs if the list is shorter) URLs. Five URL's with the highest relevance labels ($r^q_i$) are placed at the top five locations in the user feedback. Note that this is a noisy version of feedback since the linear fit cannot describe the labels exactly in this dataset.

As a baseline, a ranking SVM was trained repeatedly. In the first iteration, a random ranking was presented, the feedback ranking (as mentioned in the paragraph above) was obtained. An SVM was trained based on the pair of examples $((q_1,\by^{q_1}), (q_1,\bby^{q_1}))$.  From then on, a ranking was  presented based on the prediction from the previously trained ranking SVM. The user always returned a ranking based on the relevance labels as mentioned above; the pairs of examples were stored after every iteration. Note that training a ranking SVM after each iteration would be prohibitive since it involves cross-validating a parameter $C$ that trades-off between the margin and the slacks. Thus, we trained an SVM whenever 10\% more examples were added to the training set after the previous training. The value of the parameter $C$ was obtained via a five-fold cross-validation.\footnote{It was fixed at 100 when there were less than 50 examples.} Once a $C$ value was determined, SVM was trained on all the training examples available at that time and used it to predict rankings until the next training.

Results of this experiment are presented in Figure \ref{fig:rel-feedback}. We have provided both the mean regret as well as one standard deviation for this experiment. Since the feedback is now based on relevance labels (and not on a linear fit), the utility regret converges to a non-zero value. It can also be noticed that our preference perceptron performs significantly better compared to the SVM. It might be possible to improve the performance of the SVM by training it more often. However, this would be extremely prohibitive. For instance, the perceptron algorithm took around 30 minutes to run (which was mostly inefficient Python IO), whereas the SVM version took about 20 hours (on the same machine). 

\section{Conclusions}
We proposed a new model of online learning with preferences that is especially suitable for implicit user feedback. An efficient algorithm was proposed that provably minimizes regret. Experiments showed its effectiveness for web-search ranking. 
\begin{footnotesize}
\bibliography{hucb}

\begin{thebibliography}{1}

\bibitem{AuerCF02}
P.~Auer, N.~Cesa-Bianchi, and P.~Fischer.
\newblock Finite-time analysis of the multiarmed bandit problem.
\newblock {\em Machine Learning}, 47(2-3):235--256, 2002.

\bibitem{ACSF02}
P.~Auer, N.~Cesa-Bianchi, Y.~Freund, and R.~Schapire.
\newblock The non-stochastic multi-armed bandit problem.
\newblock {\em SIAM Journal on Computing}, 32(1):48--77, 2002.

\bibitem{olbook}
N.~Cesa-Bianchi and G.~Lugosi.
\newblock {\em Prediction, learning, and games}.
\newblock Cambridge University Press, 2006.

\bibitem{ylr11}
O.~Chapelle and Y.~Chang.
\newblock Yahoo! learning to rank challenge overview.
\newblock {\em JMLR - Proceedings Track}, 14:1--24, 2011.

\bibitem{FKM05}
A.~Flaxman, A.~T. Kalai, and H.~B. McMahan.
\newblock Online convex optimization in the bandit setting: gradient descent
  without a gradient.
\newblock In {\em SODA}, 2005.

\bibitem{Joachims/etal/07a}
T.~Joachims, L.~Granka, Bing Pan, H.~Hembrooke, F.~Radlinski, and G.~Gay.
\newblock Evaluating the accuracy of implicit feedback from clicks and query
  reformulations in web search.
\newblock {\em ACM Transactions on Information Systems (TOIS)}, 25(2), April
  2007.

\bibitem{Yue/etal/09a}
Y.~Yue, J.~Broder, R.~Kleinberg, and T.~Joachims.
\newblock The k-armed dueling bandits problem.
\newblock In {\em COLT}, 2009.

\bibitem{Yue/Joachims/09a}
Y.~Yue and T.~Joachims.
\newblock Interactively optimizing information retrieval systems as a dueling
  bandits problem.
\newblock In {\em ICML}, 2009.

\bibitem{Zink03}
M.~Zinkevich.
\newblock Online convex programming and generalized infinitesimal gradient
  ascent.
\newblock In {\em ICML}, 2003.

\end{thebibliography}
\bibliographystyle{plain}
\end{footnotesize}

\appendix

\section{Proof of theorem 1}
\label{app:proof}
\begin{proof}
First, consider the inner product of $\bw_{T+1}$ with itself. We have,
\begin{align}
\bw_{T+1}^\top\bw_{T+1} &= \bw_{T}^\top \bw_{T} + 2 \bw_T^\top (\phi(\bx_T,\bby_T) - \phi(\bx_T,\by_T)) \nnnl
&+ ( \phi(\bx_T,\bby_T) - \phi(\bx_T,\by_T) )^\top(  \phi(\bx_T,\bby_T) - \phi(\bx_T,\by_T) ) \nnnl
& \le \bw_{T}^\top\bw_{T} + 4 R^2 \nnnl
& \le 4R^2T. \nn
\end{align}
On the first line, we simply used our update rule from algorithm \ref{perceptron}. On the second line,
we used the fact that $\bw_T^\top  (\phi(\bx_T,\bby_T) - \phi(\bx_T,\by_T))  \le 0$ from the choice 
of $\by_T$ in Algorithm \ref{perceptron} and that $\| \phi(\bx_T,\bby_T) - \phi(\bx_T,\by_T)\|^2 \le 4R^2$. We obtain the last line inductively.

Further, from the update rule in algorithm \ref{perceptron}, we have,
\begin{align}
\bw_{T+1}^\top \bw^* &= \bw_{T}^\top\bw^* + \bw^{*\top}  \left( \phi(\bx_T,\bby_T) - \phi(\bx_T,\by_T) \right) \nnnl
&= \sum_{t=1}^T   \bw^{*\top} \left( \phi(\bx_t,\bby_t) -  \phi(\bx_t,\by_t) \right) \nnnl
& = \sum_{t=1}^T\left( U(\bx_t,\bby_t) - U(\bx_t,\by_t) \right).  \nn
\end{align}
We now use the fact that $\inner{\bw_{T+1}}{\bw^*} \le \| \bw^* \| \| \bw_{T+1} \|$ (Cauchy-Schwarz inequality) which implies,
$$  \sum_{t=1}^T \left( U(\bx_t,\bby_t) - U(\bx_t,\by_t) \right)   \le 2R\sqrt{T}\|\bw^*\|.$$
The above inequality, along with the $\alpha$-informative feedback (Eqn. \eqref{eq:inf-feedback-relax}) gives,
$$  \alpha \sum_{t=1}^T\left( U(\bx_t,\by^*_t) - U(\bx_t,\by_t) \right)  - \sum_{t=1}^T \xi_t \le 2R\sqrt{T}\|\bw^*\|    .$$
from which the claimed result follows.
\end{proof}

\end{document}